\documentclass[runningheads]{llncs}
\usepackage[T1]{fontenc}
\usepackage{graphicx}
\usepackage{color}
\usepackage{subcaption}

\begin{document}
\title{Applicability of memorization indicators for early spotting of overfitting while recalibrating sEMG-decoders on low sample sizes}
\titlerunning{Monitoring subject-calibration of sEMG-decoders}
%
\author{
Stephan J. Lehmler\inst{1,2}\orcidID{0000-0002-6373-4074} \and
Tobias Glasmachers\inst{3}\orcidID{0000-0003-1886-1696} \and
Ioannis Iossifidis\inst{4}\orcidID{0000-0002-9876-4396}}
\authorrunning{S. Lehmler et al.}
%
\institute{German Aerospace Center (DLR), Institute for AI Safety and Security, 
  Ulm, Germany
  \email{stephan.lehmler@dlr.de}
 \and
  Faculty of Electrical Engineering and Information Technology, Ruhr University Bochum, Bochum, Germany \and
   Institute for Neural Computation, Ruhr University Bochum, Bochum, Germany
   \and
  Institute of Computer Science, Ruhr West University of Applied Science, Bottrop, Germany
  \email{ioannis.iossifidis@hs-ruhrwest.de}
}
\maketitle              
\begin{abstract}

Deep learning models for surface electromyography (sEMG) can benefit substantially from subject-specific (re-)calibration, since no sufficiently large and diverse datasets are available to train fully generic decoders. However, for user acceptance, the number of repetitions that can realistically be collected during calibration is severely limited, which increases the risk of overfitting and, in extreme cases, can even degrade performance compared to the uncalibrated model. Classical overfitting indicators such as validation performance and regularization with early stopping are difficult to apply in this low-sample regime, as they require additional held-out data that is rarely available in practical calibration scenarios.

In this work, we investigate a recently proposed class of memorization indicators based solely on the activation statistics of rectified linear units (ReLU) in deep neural networks, which can be computed directly from training data without any extra validation set. We conduct a transfer-learning experiment on a benchmark sEMG dataset, where a convolutional neural network is first pre-trained on multiple subjects and subsequently fine-tuned on individual users using only a small number of repetitions. During calibration, we monitor both decoding performance and the activation behaviour of the last hidden layer. Our results provide first evidence that decreases in test accuracy during fine-tuning are accompanied by characteristic changes in activation rates, indicating that activation-based memorization indicators are a promising tool for early spotting of unsuccessful learning in low-sample sEMG calibration settings.


\end{abstract}

\section{Introduction}
This paper concerns the early spotting of overfitting during calibration of deep neural network (DNN)-based classification models for surface Electromyograpphy (sEMG) data via fine-tuning based transfer learning.\\
Electrical signals from muscles measured on the skin-surface, the sEMG, is one of the primary datasources \cite{lobo2014non,10.3389/fnbot.2014.00024,tam2019human} for non-invasive prediction of movement intention in humans. As such, accurate decoding of these signals via classification or regression is an important step for various applications \cite{muguro2021development,app13179546}. These include in especially medical applications, like rehabilitory exoskeletons, protheses and ortheses, but also other uses in Human-Machine-Interaction, like robotic control or gaming. Our work is motivated by the application of sEMG-classifiers for controlling upper-arm exoskeletons using small DNNs, capable to be run and calibrated locally on edge hardware.\\
In recent years, deep learning has consistently outperformed classical decoding approaches \cite{zhou2021comparison,xiong2021deep,sid2024comprehensive}. State of the art results for classifying movement intention in sEMG-signals is currently being achieved by large attention-based DNN, e.g. in \cite{shin2025electromyographybasedgesturerecognitionhierarchical,shah2025tff,lin2024continuous,10789212}.\\
However, these powerful models come with new challenges for real-life applications. As the data derived via sEMG is especially dependent on person-specific factors such as age, gender \cite{sahebjada2024effect} or electrode placement \cite{HOGREL1998305}, decoder models have  to be generally be tuned on each new user, which is not as straightforward for DNN, with their large parameter space, as for simpler machine learning models. To ease user adoption, a quick and effortless calibration process is required, which means that one can realistically only demand a few sample repetitions from each new user to be used for the calibration.  Here, transfer learning has been shown to be beneficial to improve calibration speed and accuracy \cite{fratti2024multi,fan2023improving,lehmler2022deep}. Still, there is the potential issue of overfitting during calibration, especially when only few movement repetitions are available for calibration. Previous work has shown how this can be an issue in transfer learning for DNN-based sEMG-decoders, and can even reduce within subject performance compared to uncalibrated models, as models would overfit by assumedly memorizing the known repetitions and perform worse on unseen repetitions of the same user \cite{lehmler2022deep}.\\
This indicates the importance of closely monitoring model performance metrics to verify a satisfactory outcome of the transfer learning process. Classical approaches for this, such as early stopping, rely on an additional validation dataset and are therefore difficult to implement for user calibration of sEMG decoders. This validation data would either demand new users to perform additional repetitions and prolonging the calibration process or have to be taken from the same repetition, which in the past was shown to be insufficient for stopping overfitting \cite{lehmler2022deep}. The calibration process would benefit from simple indicators for overfitting or other forms of unsuccesful learning, that can be evaluated reliably on few samples from the existing training data.\\
Recent work \cite{lehmler2024understanding,lehmler2025} has introduced memorization indicators based on a stochastic process model of the activation behaviour of neurons with rectified linear units (ReLU). The work proposes simple distribution metrics representing the activation rate of neurons in the layer used for training. When comparing these metrics for generalizing image classifiers with the same models forced to memorize via label randomization, consistent differences where shown. First and foremost, the memorizing models showed a significantly lower activation rate.\\
If this connection holds also in the context of other types of data and overfitting models, activation rate based metrics can be a promising candidate for a quantitative monitoring of transfer learning based calibration in low sample settings. While the indicators have originally been introduced for discerning generalization and memorization, it has been argued \cite{lehmler2025} that the relationship between decreasing activation rates and memorization is caused by the network learning (or memorizing) increasingly 'rarer' features of the data. As such, the underlying model of memorization is, compared to other works \cite{usynin2024memorisation}, less strictly construed around the idea of memorization as remembering singular data points. Therefore, the proposed indicators should be applicable to the overfitting on small sEMG samples, where it is less clear-cut if the overfitting is likely caused by memorization of circumstantial features of the fine-tuning data instead of by strict memorization of singular samples. 
In this paper, we investigate whether or not this apparent relationship between the distribution of activation rates and learning outcomes holds for sEMG signals and classifiers. In addition to the previously introduced indicators, we add further metrics based on activation rate statistics. We show promising first results, indicating that a drop in test accuracy during transfer learning does come with a noticeable reduction of activation rates. The results are stable in a realistic experiment with data from 10 subjects. 

\section{Data \& Experiment}
We implement a small transfer learning experiment on a representative sEMG benchmark dataset, utilizing a pre-training phase, followed by a subject-specific calibration or fine-tuning phase. During this second phase, we record activation characteristics in the last hidden layer of the DNN.\\
For our experiment, we utilize an existing sEMG benchmark-dataset, the well-known NinaPro database \cite{db2}, particular Database 2, which measures 49 movement classes (+rest) and consists of 40 subjects in total, out of which we use data from the first 30 in our experiment. All data has been recorded on healthy (intact limbs) subjects using 12 Delsys Trigno electrodes. For each of the 49 classes (grasping, finger and various functional movements) have been recorded 6 successive repetitions (4/2 Training/Test-Split).\\
In our experiment, we take the raw sEMG measurements as input with only minimum pre-processing. The only pre-processing steps are a data normalization based on the training data and a windowing step, splitting the sEMG timeseries in windows of 50-samples (20ms) each with an overlap of 10-samples. Data access and processing has been implemented using the python-package LibEMG \cite{10214558}. 
For the pre-training, we use subject 1-20, while the transfer-learning phase uses subject 21-30.\\
\\
As deep-learning model, we chose a comparatively simple 1D-Convolutional Network. It takes the 12x50 (20ms@2kHz of 12 sEMG channels) as inputs, processes them using 3 1D-CNN-blocks with fixed kernel-size (5) decreasing numbers of filter (256, 128,64) Batchnorm and ReLU-activation after which 3 fully connected ReLU-layers (width: 1216, 608, 256) compress the CNN-output. The classification layer had a width of 49 and softmax activation. For training, we use Adam with CrossEntropyLoss and a fixed learning rate of 0.001 for 100 epochs using a batchsize of 64.\\
While this model is obviously nowhere near the size and complexity of state-of-the-art (SOTA) models in the field, it does show a realistic model architecture for our application area. As we want models capable to run inference in real-time as well as acceptable calibration time with local fine-tuning on edge-devices without GPU or AI-acceleration, larger models are currently out-of-scope. Even with the chosen small-sized architecture, better performance than achieved might be possible with further hyperparameter tuning. As the aim of our experiment is however not focused on final model performance, but having a realistic model architecture for monitoring deviant behaviour during calibration, we deemed the trained model to be adequate for addressing the research question.\\
\\
After training this initial model, the following calibration phase is the main focus of the experiment.
During this phase, we first take the data generated by each of the subjects 21-30 and evaluate the pretrained models performance on their respective test data. Afterwards we fine-tune the model on a number of repetitions of the given subjects for 50 epochs. After each epoch, we log information about the performance on the train- and test-data, as well as on the activations of the last fully-connected hidden layer (size: 256 neurons).\\
In trying to induce a degradation in the test-performance (e.g. induce overfitting), we varied two experimental conditions. The first is the number of repetitions used for the calibration, as smaller sample sizes have resulted in clearly overfitting models in earlier experiments \cite{lehmler2022deep}. We trained on a range of 1 to 4 repetitions, with one repetition containing one example per movement class each. The second experimental condition was the number of trainable parameters. In one experimental condition, we froze all but the last two layers (calibration-layers) of our model. In the second condition, we retrained all the parameters of the model, under the assumption, that the higher parameter space would more easily lead the models to overfit. The hyperparameter during this phase were identical to the pre-training phase. \\
\\
The main parameter of interest in this paper is the mentioned activation rate of a neuron. The activation rate is being calculated exclusively for the last hidden layer of the model (in accordance to experimental setups in \cite{lehmler2024understanding}).
To estimate the activation rate of one ReLU-neuron, the number of times its output is higher than 0 is counted and divided by the number of samples processed.  \\
It is based only on the first batch (64 samples) of each epoch during the fine-tuning phase, so it is recording the behaviour of neurons on the training data. This is suitable to the initially stated need for indicators that don't need additional validation datasets.

\section{Results}

\begin{figure} 
    \centering
    \includegraphics[width=\linewidth]{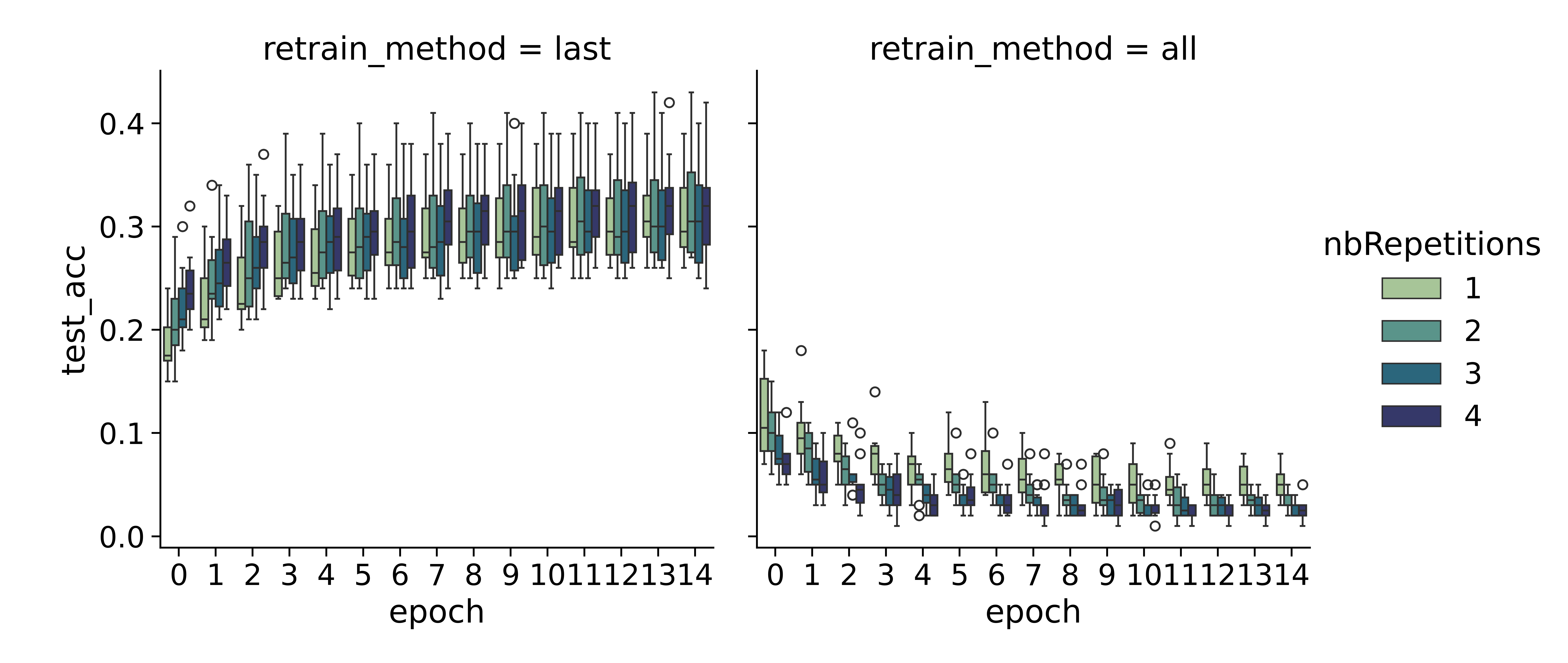}
    \caption{Distribution of test accuracy during the first 15 epochs of calibration over 10 unseen subjects with a varied number of repetitions used for fine-tuning.}
\label{fig:performance}
\end{figure}

After the initial training on 20 subjects, the model described above reached a train accuracy of 0.52 and a test accuracy of 0.34. Without calibration however, on the unknown subjects 21-30, the same models accuracy was only 0.065 on average (std: 0.017). Initial performance of the pretrained network on data from new users was therefore only slightly above chance (at 49 classes $\approx$ 0.02), highlighting the importance of retraining.\\
Figure \ref{fig:performance} shows the model performance on the test set (repetition 5 \& 6) of each new subject during the initial 15 epochs of fine-tuning. We see a clear difference in the learning result of both retraining methods. Only the model with the last two layers used unfrozen as calibration layer generalizes to unseen repetitions of the same user, while the highly overparametrized model with all layers being retrained does not generalize at all. We will therefore use the model with the retrain method 'last' and the model retrained using 'all' layers serve therefore as examples of successful and unsuccessful fine-tuning respectively. It should be noted that the later does not necessarily overfit, as it also does not learn the training data successfully.\\
The number of repetitions used for fine-tuning does not appear to have any relevant effect on final performance in our experiments.
In the following, we will look first at the relationship of changes in Mean Activation Rate (MAR) and learning performance and later see how MAR and related measures change online during fine-tuning.

\subsection{Relationship between Activation Rate and Transfer-Learning Performance}
\begin{figure} 
    \centering
    \includegraphics[width=\linewidth]{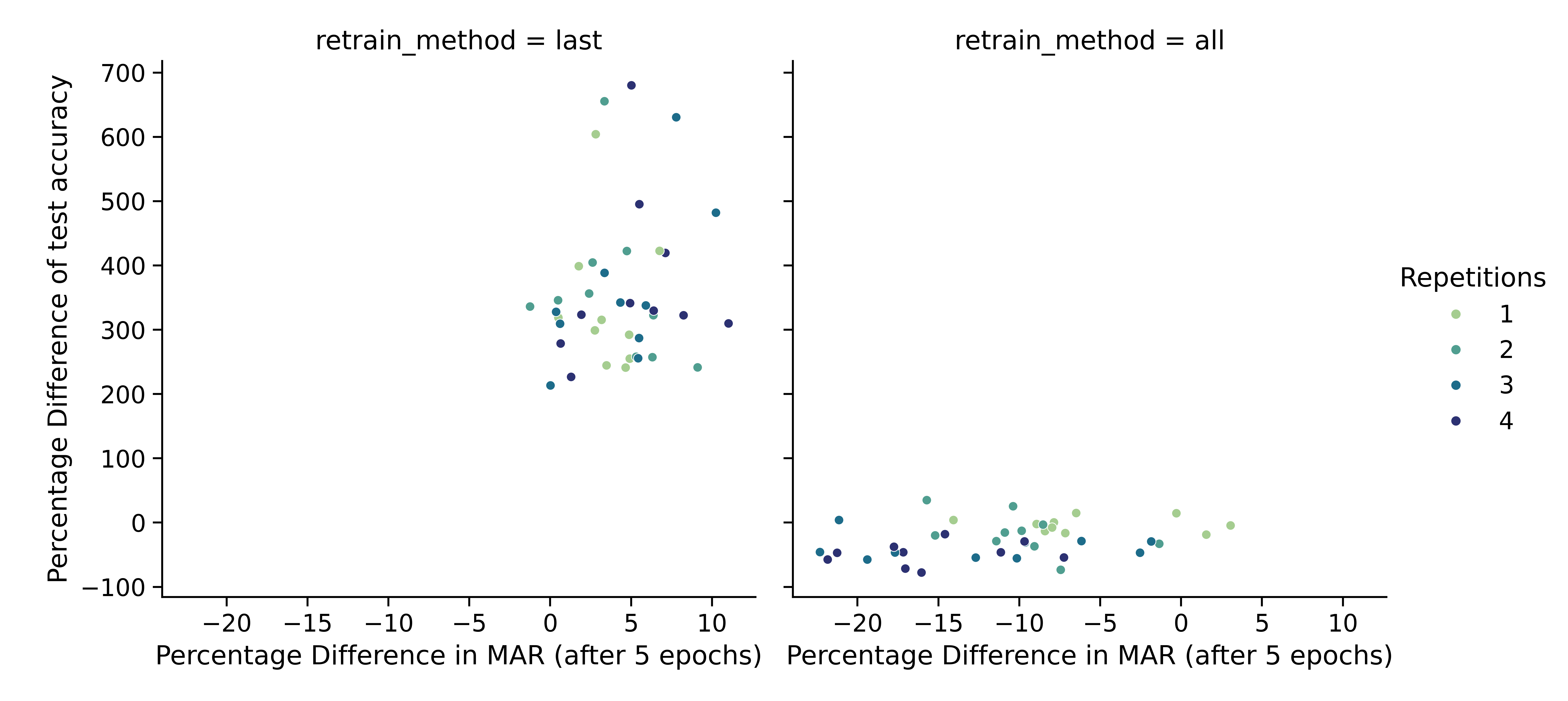}
    \caption{Relationship between percentage changes in test accuracy and Mean Activation Rate (MAR) early on during the calibration (after 5 epochs). The X-Axis shows how much MAR (evaluated on 64 samples) changed after 5 epochs compared to the initial state, the Y-Axis shows how much the test accuracy improved relative to pre-training conditions. Colour represents the number of iterations used for fine-tuning.\\
The learning and the non-learning model can be clearly separated by the direction and magnitude of changes in MAR early on. }
\label{fig:scatter}
\end{figure}

\begin{figure}
\begin{subfigure}{.5\linewidth}
  \centering
  \includegraphics[width=.9\linewidth]{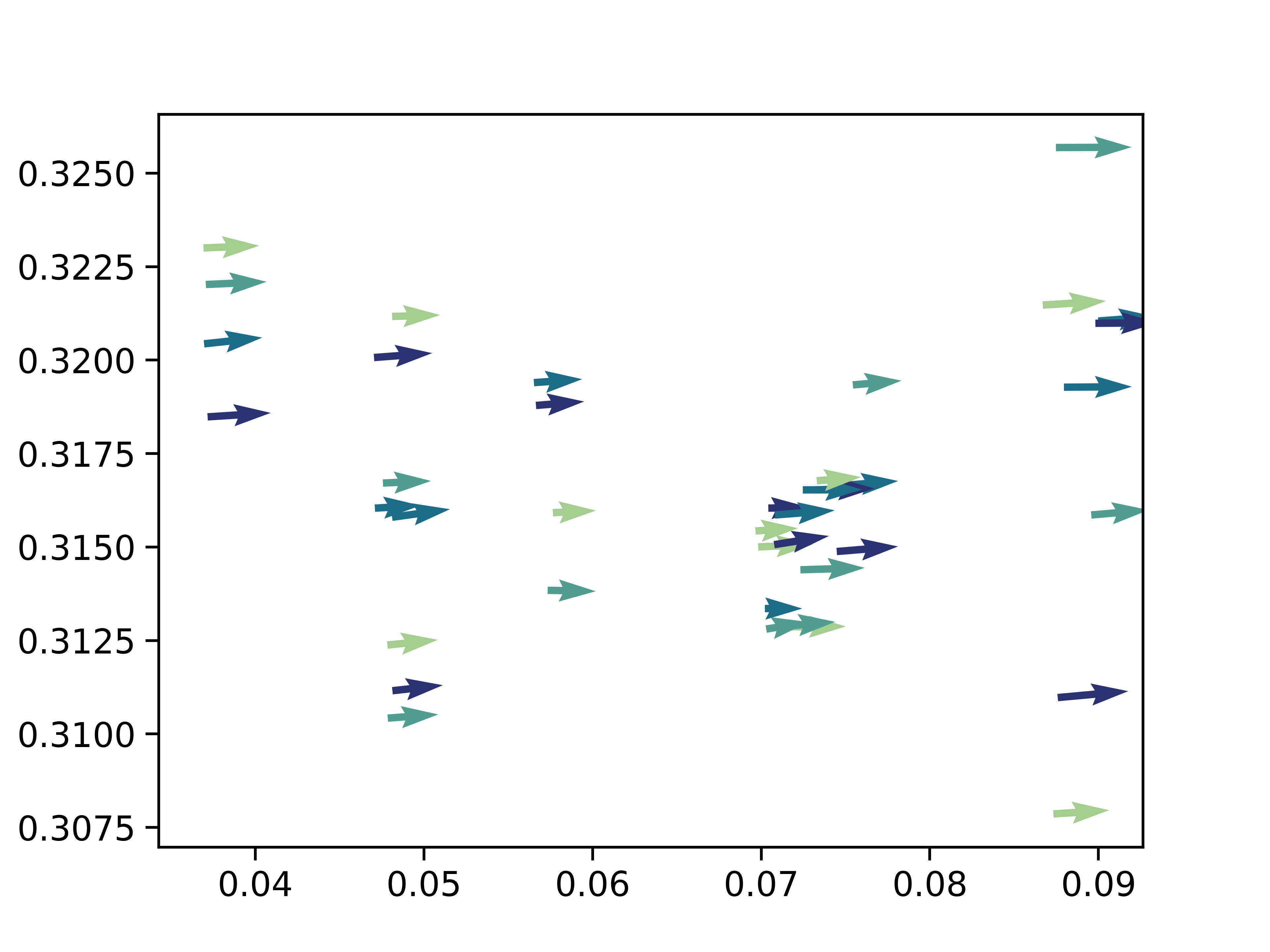}
  \caption{After 5 Epochs trained on \textbf{last} layers}
  \label{fig:sfig1}
\end{subfigure}%
\begin{subfigure}{.5\linewidth}
  \centering
  \includegraphics[width=.9\linewidth]{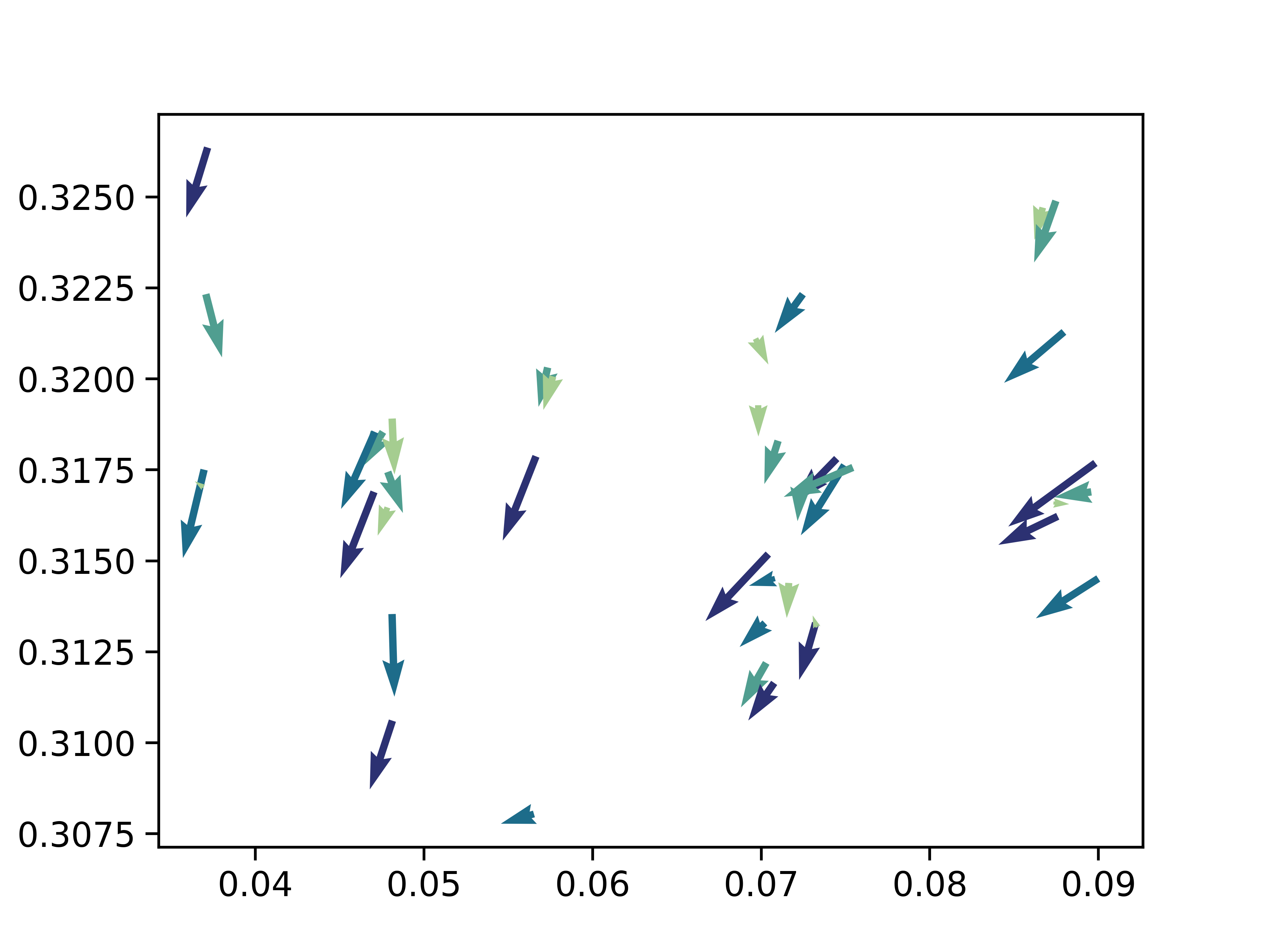}
  \caption{After 5 Epochs trained on \textbf{all} layers}
  \label{fig:sfig2}
\end{subfigure}
\begin{subfigure}{.5\linewidth}
  \centering
  \includegraphics[width=.9\linewidth]{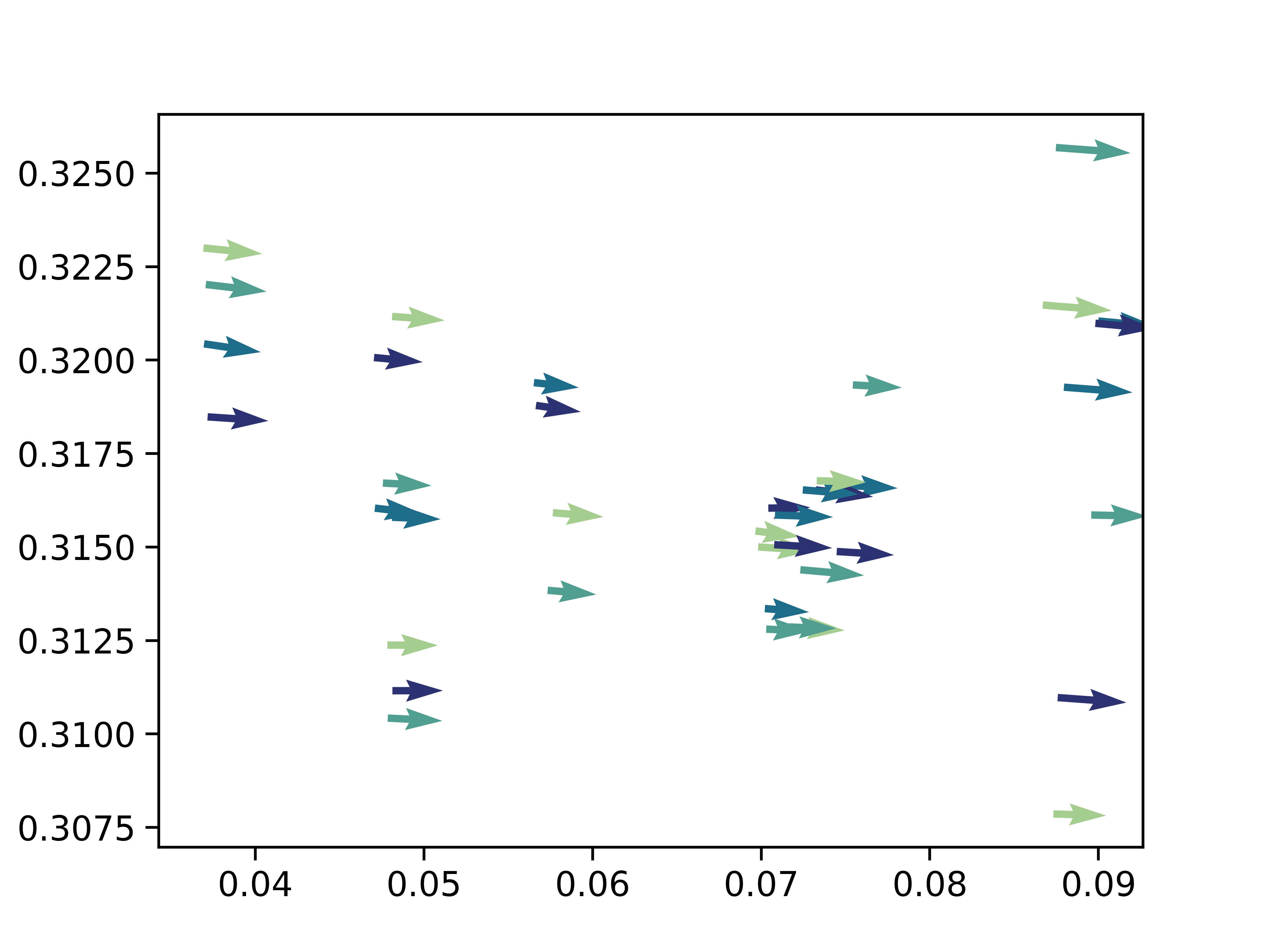}
  \caption{After 50 Epochs trained on \textbf{last} layers}
  \label{fig:sfig1}
\end{subfigure}%
\begin{subfigure}{.5\linewidth}
  \centering
  \includegraphics[width=.9\linewidth]{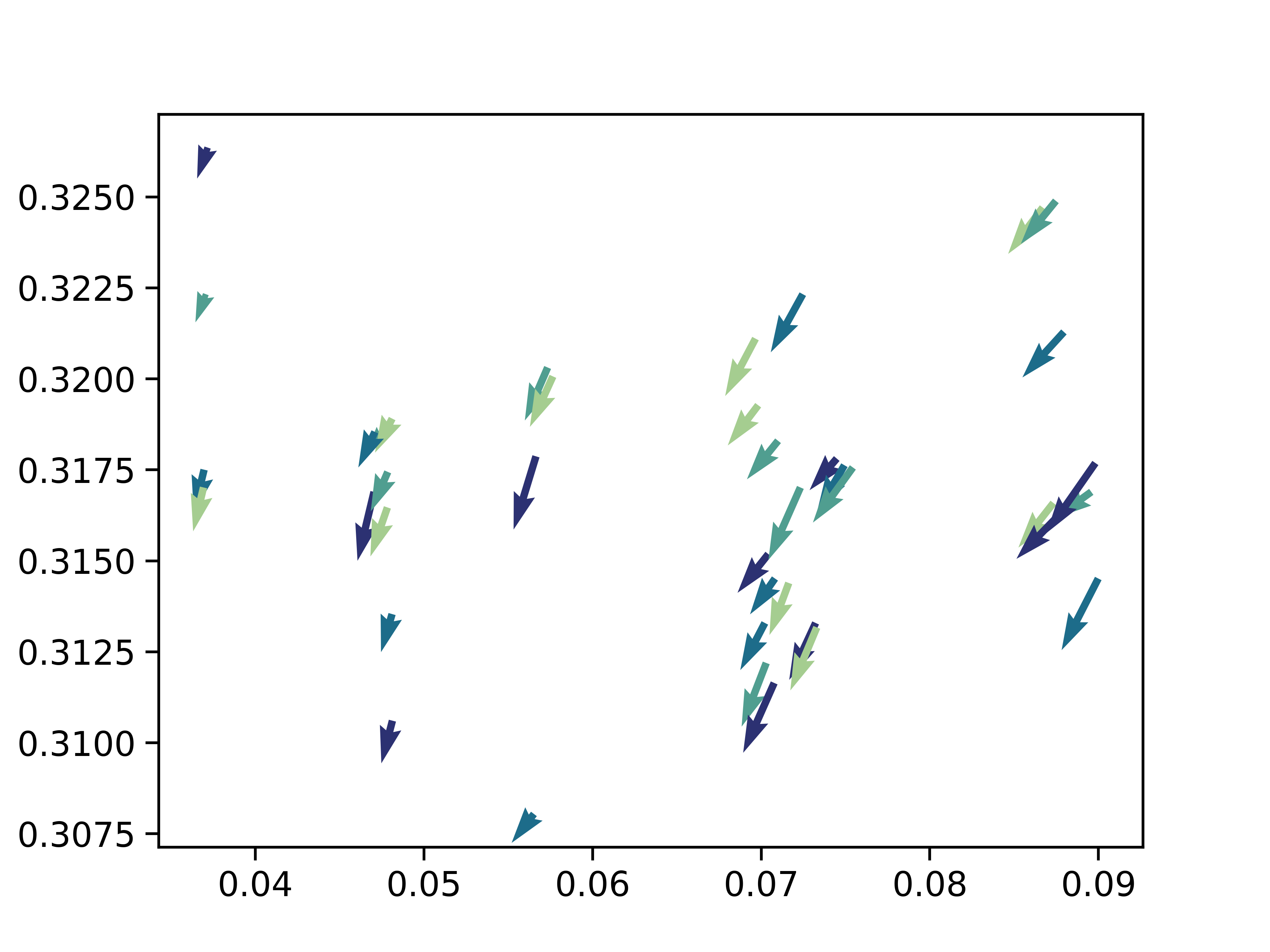}
  \caption{After 50 Epochs trained on \textbf{all} layers}
  \label{fig:sfig2}
\end{subfigure}
\caption{Quiver plots of changes during the experiments. The X-Axis shows the initial test-set performance (accuracy) before fine-tuning. The Y-Axis shows the Mean Activation Rate (MAR) of the 256 neurons of the last layer, calculated on 64 training samples. Each arrow represents one calibration experiment, its starting point is the metrics at the beginning of the experiment. Its direction showed a \textit{scaled} version of the direction both variables moved during the experiment (the tip does \textit{not} show the final values). Colour represents the number of iterations used for fine-tuning.\\
Subplots clearly show a difference between successfully and unsuccessfully calibrated models. We see in (a) an upward-trend in model performance after 5 epochs, while MAR stays roughly the same. In (b), we see how MAR and model performance decrease simultaneous. Only after 50 epochs, we also notice as slight drop in MAR despite increased model accuracy (c). This decrease is however way less pronounced as in the cases of unsuccessful learning shown in (b) and (d).}
\label{fig:quiver}
\end{figure}

We are interested in whether the absolute value or change of Mean Activation Rate (MAR) of the final hidden layer observed on a small batch of the training data (64 samples) is indicative of model performance. As the same pretrained model is used, the initial (pre-calibration) MAR-values for all subjects are close together (mean: 0.317, std: 0.004, min: 0.308, max: 0.326), but statistically not identical, as the Kruskal-Wallis test for equality in means over multiple groups (subjects) does show insignificant results (H-value: 8.643, p-value: 0.28). The raw value of MAR seems therefore to be to at least some degree be dependent on the data used for the evaluation. But as can be seen in figure \ref{fig:scatter}, both learning methods can be separated rather well by their changes in MAR. After only 5 epochs, the  model showing improvement on the test data (retraining method = last), shows for nearly all subjects and sample sizes a small positive change. The worse performing models (with negative changes of accuracy) do overwhelmingly tend to decreasing MAR, with an overall higher change, but wider variability in reduction. We however don't observe any clear functional (linear or otherwise) relationship that would indicate a quantitative predictability based on changes in MAR. 
We also notice that this general tendency towards large negative changes in MAR is stable for later epochs (figure \ref{fig:quiver}) of the models with decreasing accuracy on the test data. With ongoing training, we notice a developing negative slope for the learning model too (figure \ref{fig:quiver} c), which might indicate that under this method too, an overfitting starts to develop. Indeed, for these models, the accuracy gap between training and testing starts relatively low (0.3445 train vs.0.2895 test after 5 epochs), but increases later (0.439 train vs.0.33725 test after 50 epochs). If MAR is indeed related to memorization, the slight decrease in MAR might actually be caused by this increasing generalization gap.

\subsection{Selected Indicators as online predictors}
\begin{figure} 
    \centering
    \includegraphics[width=\linewidth, height=600pt]{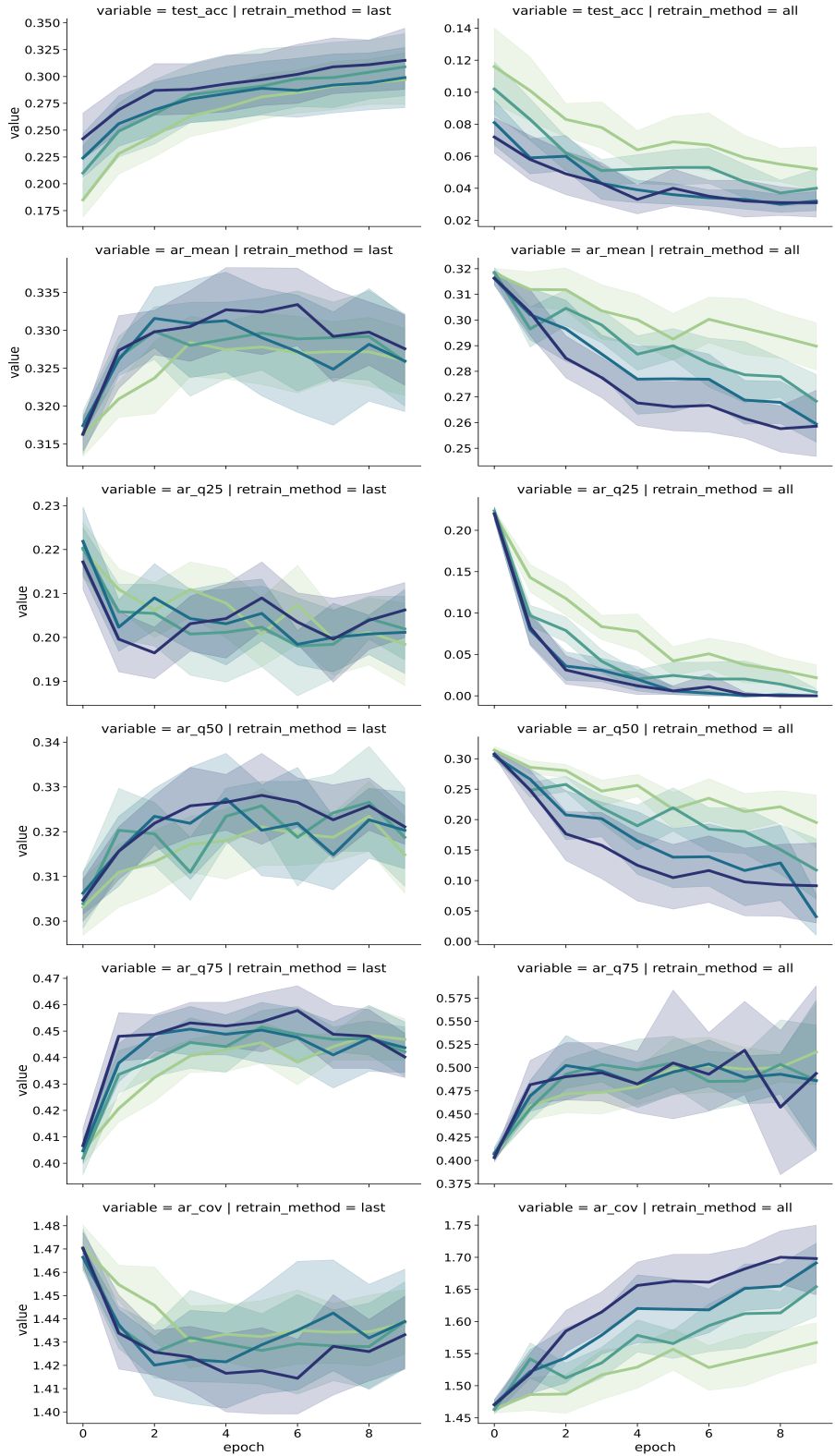}
    \caption{Changes in multiple Activation Rate (AR) based metrics during calibration for the first 10 epochs. The first row shows the accuracy on the test data, row 2-6 measures of the distribution of AR in the last hidden layer observed on the training data. The Y-Axis is differently scaled for each plot.}
    \label{fig:timeseries}
\end{figure}

In the previous section, we have shown how MAR might proof itself as a useful measure to distinguish overfitting models in realistic sEMG applications. As initially stated, we are interested in using AR-based indicators as a quantitative monitoring metric for monitoring the calibration/learning progress. For this we need to understand the dynamics of these indicators, which can be seen in figure \ref{fig:timeseries}\\
We want to capture more details about the distribution of activations in the observed hidden layer. For this we add to the already discussed mean AR, the quantile-distribution (25\%, median, 75\%) and the coefficient of variation. The later has been already used in \cite{lehmler2024understanding}, where an increased variance in AR has been reported for memorizing models.We observe a similarly an increase during overfitting. In our case, it continuously rises for the badly performing fine-tuning method. Initially starts to drop in the well performing model, but slowly starts to increase as train and test performance start to drift apart. For the other distributional indicators newly introduced in this paper, we see that the AR-quantiles do add some details of distributional changes that are getting lost if only the MAR is being considered.

\section{Discussion}
\subsection{Limitation }
The presented study concentrates on compact neural architectures and reduced experimental complexity to ensure feasibility for real-time, on-device calibration scenarios. This focus enables precise control over the learning dynamics and allows a clear assessment of activation-based indicators under realistic constraints relevant for embedded systems. While these design choices inherently limit absolute model performance compared to SOTA approaches, they strengthen the interpretability and practical relevance of the observed behaviour for resource-constrained applications.

The transfer-learning experiments were intentionally conducted under challenging, low-sample conditions to target settings in which overfitting is particularly likely to emerge. As a consequence, calibration success varied across subjects, highlighting the difficulty of balancing adaptability and robustness in small data regimes. Rather than being a drawback, this variability provides valuable insight into how sensitively memorization indicators respond to different manifestations and severities of overfitting in practice.

Furthermore, the characteristic scale and expected range of memorization activation rate (MAR) values may depend on dataset-specific properties such as the number of classes, recording setup, and sampling frequency, as is typical for sEMG benchmarks like NinaPro DB2. This work therefore does not propose universal MAR thresholds, but instead demonstrates the feasibility of using activation statistics as a relative, within-experiment indicator. Nonetheless, stable qualitative behaviour of MAR and related indicators has been demonstrated. The discussed effects are in concordance with existing research on other datasets and shown to allow insights into the training dynamics during the subject-specific calibration of an sEMG-decoder. The combination of multiple more detailed measures of activation statistics in addition to MAR might present a fruitful research direction. Future research should extend the analysis to a broader spectrum of datasets, architectures, and modalities to derive more standardized reference ranges and to investigate how MAR behaves in larger and deeper networks. This would directly support the development of more general calibration-monitoring schemes that leverage activation-based metrics as part of adaptive, closed-loop training strategies.

\subsection{Conclusion}
In this study, we investigated activation-based memorization indicators as a tool for early detection of overfitting during subject-specific calibration of DNN-based sEMG decoders in low-sample transfer-learning settings. Using a compact convolutional architecture on the NinaPro DB2 benchmark and minimal preprocessing based on windowed raw sEMG, we observed that deteriorating test performance during fine-tuning is often accompanied by a reduction in the activation rate of ReLU neurons in the last hidden layer. This suggests that simple activation statistics can capture aspects of memorization dynamics that are not directly visible from training accuracy alone.

Because the investigated indicators operate solely on training data and do not require a separate validation set, it is particularly well suited for embedded, user-centric calibration scenarios, where additional repetitions are costly or impractical. Within this constrained but practically relevant setting, our results provide first empirical evidence that mean activation rate (MAR) can serve as a lightweight, online-compatible proxy for impending overfitting. At the same time, our findings underline that MAR should currently be interpreted as a relative measure whose absolute scale depends on the dataset and architecture. This work also investigated, for the first time, other statistical metrics besides the mean of the activation rate distribution, which should be investigated further.

Future work will extend this analysis to larger and deeper networks, alternative activation functions, and additional sEMG and non-sEMG datasets, with the goal of deriving more standardized MAR reference ranges and understanding cross-domain invariances. In the longer term, we envision activation-based memorization indicators being integrated into adaptive calibration pipelines, where they could trigger early stopping, selective layer freezing, or dynamic regularization schedules. Such mechanisms would support more robust, data-efficient fine-tuning of sEMG-decoders on edge hardware, thereby facilitating reliable deployment of myoelectric control in real-world human–machine interaction scenarios.

\begin{credits}

\subsubsection{\discintname}
The authors have no competing interests to declare that are relevant to the content of this article. 
\end{credits}
%
%
%
\bibliographystyle{splncs04}
\bibliography{references}

\end{document}